\documentclass[letterpaper, 10 pt, journal, twoside]{ieeetran} 
\IEEEoverridecommandlockouts                              % This command is only needed if 
\usepackage{graphics} % for pdf, bitmapped graphics files
\usepackage{epsfig} % for postscript graphics files
\usepackage{times} % assumes new font selection scheme installed
\usepackage{amsmath} % assumes amsmath package installed
\usepackage{amssymb}  % assumes amsmath package installed
\usepackage{subfig}
\usepackage{multirow}
\makeatletter
\let\NAT@parse\undefined
\def\ps@IEEEtitlepagestyle{% default title page headers, no footers
\def\@oddfoot{\scriptsize 
~\copyright2022 IEEE. Personal use is permitted, but republication/redistribution requires IEEE permission.
see https://www.ieee.org/publications/rights/index.html for more information.
\hfill\thepage}%
}
\makeatother

\usepackage[pagebackref=true,breaklinks=true,colorlinks,bookmarks=false]{hyperref}

\newcommand{\tablestyle}[2]{\setlength{\tabcolsep}{#1}\renewcommand{\arraystretch}{#2}\centering\small}

\newcommand{\bd}[1]{\textbf{#1}}
\newlength\savewidth\newcommand\shline{\noalign{\global\savewidth\arrayrulewidth
  \global\arrayrulewidth 1pt}\hline\noalign{\global\arrayrulewidth\savewidth}}

\title{
BIMS-PU: Bi-Directional and Multi-Scale Point Cloud Upsampling
}

\author{Yechao Bai$^{1}$, Xiaogang Wang$^{1}$, Marcelo H. Ang Jr$^{1}$ and Daniela Rus$^{2}$% <-this % stops a space
\thanks{Manuscript received: February, 22, 2022; Revised May, 18, 2022; Accepted June, 9, 2022.}%Use only for final RAL version
\thanks{This paper was recommended for publication by Editor Cesar Cadena upon evaluation of the Associate Editor and Reviewers' comments. 
This work was supported by the National Research Foundation, Prime Minister's Office, Singapore, under its CREATE program, Singapore-MIT Alliance for Research and Technology (SMART) Future Urban Mobility (FM) IRG. (\textit{Yechao Bai and Xiaogang Wang are co-first authors.})(\textit{Corresponding author: Yechao Bai.})} %Use only for final RAL version
\thanks{$^{1}$Yechao Bai, Xiaogang Wang and Marcelo H. Ang Jr are with the Department of Mechanical Engineering, National University of Singapore, Singapore.
        {\tt\footnotesize \{yechao.bai, xiaogangw\}@u.nus.edu, mpeangh@nus.edu.sg}
        }%
\thanks{$^{2}$Daniela Rus is with the Massachusetts Institute of Technology, Cambridge, MA, USA.
        {\tt\footnotesize rus@csail.mit.edu}}%
}

\begin{document}

\maketitle
% \thispagestyle{empty}
% \pagestyle{empty}

%%%%%%%%%%%%%%%%%%%%%%%%%%%%%%%%%%%%%%%%%%%%%%%%%%%%%%%%%%%%%%%%%%%%%%%%%%%%%%%%
\begin{abstract}
The learning and aggregation of multi-scale features are essential in empowering neural networks to capture the fine-grained geometric details in the point cloud upsampling task.
Most existing approaches extract multi-scale features from a point cloud of a fixed resolution, hence obtain only a limited level of details. 
Though an existing approach aggregates a feature hierarchy of different resolutions from a cascade of upsampling sub-network, the training is complex with expensive computation.
To address these issues, we construct a new point cloud upsampling pipeline called BIMS-PU that integrates the feature pyramid architecture with a bi-directional up and downsampling path.
Specifically, we decompose the up/downsampling procedure into several up/downsampling sub-steps by breaking the target sampling factor into smaller factors. 
The multi-scale features are naturally produced in a parallel manner and aggregated using a fast feature fusion method.
Supervision signal is simultaneously applied to all upsampled point clouds of different scales.
Moreover, we formulate a residual block to ease the training of our model.
Extensive quantitative and qualitative experiments on different datasets show that our method achieves superior results to state-of-the-art approaches.
Last but not least, we demonstrate that point cloud upsampling can improve robot perception by ameliorating the 3D data quality.
\end{abstract}

% \begin{keywords}
% Deep Learning for Visual Perception, Point Cloud Upsampling.
% \end{keywords}

% Keywords appear just beneath the abstract. Use only for final RAL version.  

\bstctlcite{IEEEexample:BSTcontrol}

%%%%%%%%%%%%%%%%%%%%%%%%%%%%%%%%%%%%%%%%%%%%%%%%%%%%%%%%%%%%%%%%%%%%%%%%%%%%%%%%
\section{INTRODUCTION}
\IEEEPARstart{T}{he} importance of 3D data has become evident in applications like autonomous driving, robotics, medical imaging, etc.
Recent studies \cite{pointnet, pointnet++, point_generation, point_transformer, cascaded} have shown that point cloud is a compact and efficient 3D representation. 
However, real-scanned point clouds produced by depth camera and LiDAR are often sparse, noisy, and irregular \cite{irregular1, irregular2, irregular3}. 
As point cloud upsampling can improve the quality of real-scanned data by increasing the point density and uniformity, it has drawn increasing attention in computer vision and robotics community.
% ####################################################################
\begin{figure}[t]
\begin{center}
  \includegraphics[width=1.0\linewidth]{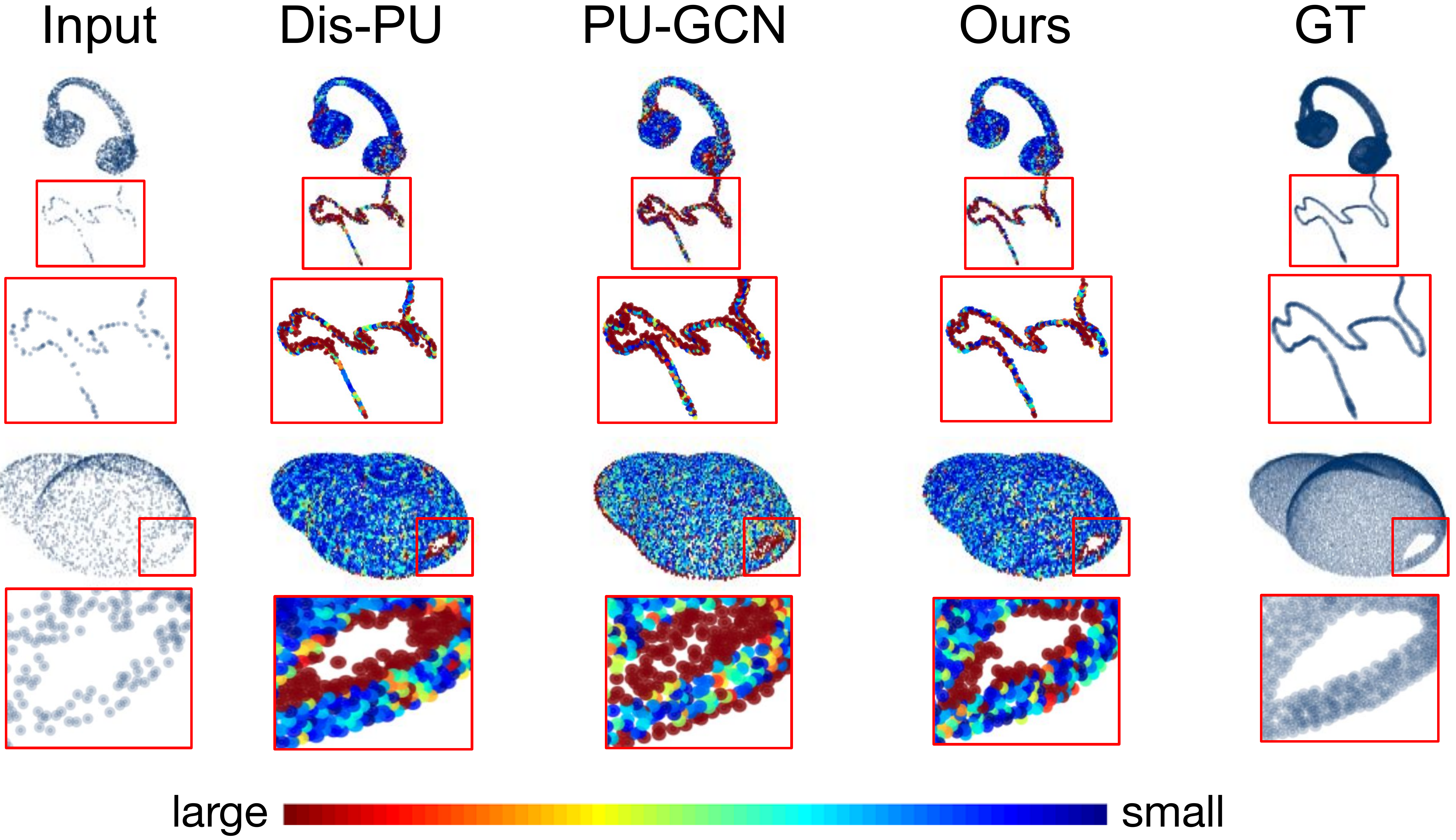}
\end{center}
\vspace{-3mm}
  \caption{
  % Comparison between outputs of our model with the result from state-of-art methods. 
  Qualitative Comparison between our model and state-of-art methods. 
  The color indicates the nearest distance of each output point to the ground truth surface.
  The result demonstrates that our approach has distinct advantages at challenging places like intersections, the narrow gap between two surfaces, and slender objects. 
  We credited it to the effective multi-scale feature fusion, which enables the model to leverage local and global contextual information.
  }
\vspace{-7mm}
\label{fig:teaser}
\end{figure}
% ####################################################################
The upsampled point cloud has to preserve the geometric detail of the underlying surface. 
To this end, several state-of-the-art point cloud upsampling methods~\cite{pu-net,mpu,pu-gcn,dis-pu} exploit the multi-scale features of the point cloud.
PU-Net~\cite{pu-net} progressively increases the ball queries of each point to extract multi-scale local region features and concatenates them to generate the upsampled features.
Following a similar principle as PU-Net, PU-GCN~\cite{pu-gcn} designs an Inception DenseGCN module, which has parallel DenseGCN branches of different receptive fields, to encode multi-scale context of point clouds. 
However, the level of geometric detail in the aggregated multi-scale feature is limited as the input resolution is fixed. 
To get fine-grained details, MPU~\cite{mpu} breaks the upsampling network into successive subnets to progressively upsample the point clouds. 
Although MPU preserves better details, its back-and-forth conversion between high-dimension feature space and 3D spatial space leads to increased computation complexity and training difficulty.
% #################################################################
\begin{figure*}
\begin{center}
  \includegraphics[width=0.90\linewidth]{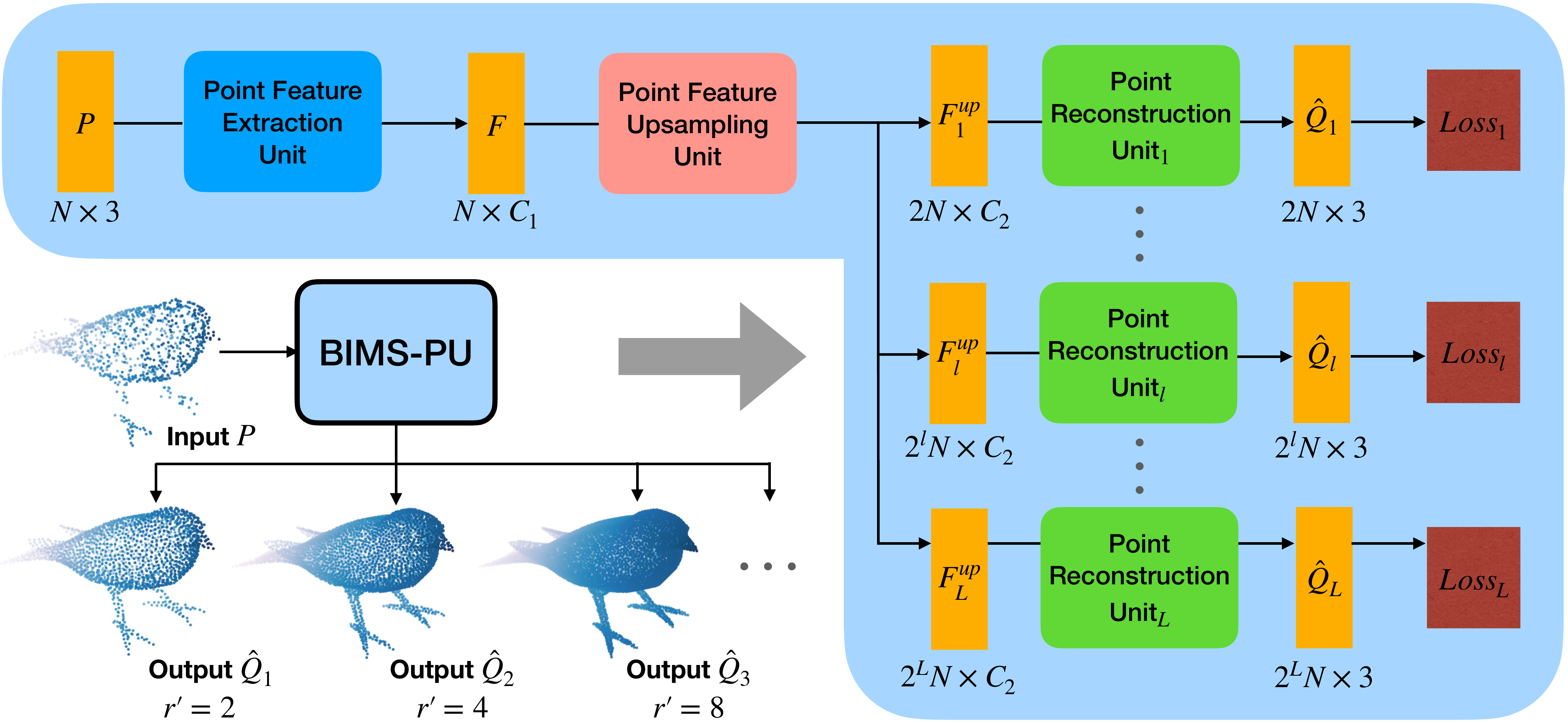}
\end{center}\vspace{-3mm}
  \caption{Overview of the bi-directional multi-scale point cloud upsampling network (BIMS-PU).
  The architecture of our model has $L$ pathways to upsample a given sparse point cloud $P$ to increasingly denser outputs $Q_1, \dots Q_L$ at the same time, as depicted in the lower left of the figure. Learning a multi-scale point cloud upsampling task enables our model to capture rich geometric details of different levels of resolution.
  }
  \vspace{-6mm}
\label{fig:network_overview}
\end{figure*}
% #################################################################
In this work, we adapt the feature pyramid architecture \cite{fpn,panet, nas-fpn, efficient-det} for the point cloud upsampling task and construct a bi-directional and multi-scale upsampling module.
Concretely, instead of obtaining the multi-scale feature directly from a feature extractor, our method generates multi-scale point cloud features from a bi-directional up and downsampling pathway inspired by the back projection mechanism \cite{back_projection} developed for image super-resolution \cite{bilateral_bp, recurrent_bp, deep_bp, feedback_bp}. 
The back projection mechanism uses an iterative up and downsampling procedure to minimize reconstruction errors.
Our method decomposes the upsampling/downsampling procedure into sub-steps with a smaller ratio in feature space and generates the multi-scale upsampled point cloud features in a parallel manner. 
The goal is to decrease the optimization difficulty via decomposing the task into multiple simpler sub-tasks \cite{acfnet}.
Next, we leverage a fast weighted feature fusion method to aggregate the resultant multi-scale features.
Subsequently, upsampled point cloud features of each scale get reconstructed into 3D point clouds. 
Supervision signals are applied simultaneously at each output scale. 
The multi-scale supervision guides the network's training to fuse multi-scale features to be more discriminative.
To enable the model to learn complex mapping with additional height and width, we formulate a simple yet highly effective residual block based on the residual learning concept \cite{resnet} to expand/squeeze the number of channels of a point feature during feature expansion and point reconstruction.
The quantitative and qualitative results show that the bi-directional multi-scale up/downsampling pathway improves the fine-grained geometric details of the upsampled point cloud. 
This is because it enables the up/downsampling operators to be trained with features of different resolutions produced by multi-scale features fusion.
Lastly, to verify the value of point cloud upsampling to the robotics community, we design an experiment to show that upsampling is beneficial to point cloud classification, a fundamental robot perception task.
In summary, our contributions are:
\begin{itemize}
    \item Design a bi-directional multi-scale upsampling module;
    \item Propose training with multi-scale supervision to facilitate the multi-scale feature fusion to produce more discriminative features;
    \item Conduct extensive quantitative and qualitative experiments on synthetic and real-world datasets to show that our method achieves superior results to state-of-the-art approaches;
    \item Design an experiment to verify the value of point cloud upsampling to robot perception.
\end{itemize}

\section{RELATED WORKS}
\noindent\textbf{Learning-based point cloud upsampling.}
PointNet \cite{pointnet} and the multi-scale variant PointNet++ \cite{pointnet++} propose networks that directly consume point cloud for several 3D recognition tasks. 
Based on PointNet++,
PU-Net \cite{pu-net} designs a point-based network for the point cloud upsampling task. It adopts the hierarchical feature learning mechanism \cite{pointnet++} for feature extraction and upsamples point cloud at patch-level. 
EC-Net \cite{ec-net} designs an edge-aware network and a joint loss to deliberately improve the consolidation near the edge. However, it requires expensive edge annotation for training.
To generate outputs of large upsampling factor,
MPU \cite{mpu} progressively upsampled the point cloud to different levels of resolution with a cascade of sub-networks.
Unlike previous encoder-decoder networks,
PU-GAN \cite{pu-gan} incorporates the adversarial training concept into a point upsampling network. 
It uses a self-attention unit to leverage the long-range context dependencies in the upsampling module.
The geometric-centric network, PUGeo-Net \cite{pugeo}, explicitly learns the first and second fundamental forms for point cloud upsampling. %
However, it requires the normals of points as a supervision signal, which is not directly available in a real-scanned point cloud.
PU-GCN \cite{pu-gcn} focuses on improving the upsampling module and the feature extraction module for point cloud upsampling. 
It integrates the graphical convolutional network into the upsampling module and designs an Inception-based feature extraction module. 
The recently proposed Dis-PU \cite{dis-pu} disentangle the point upsampling tasks into dense point generation and point spatial refinement.
To achieve this, they design a network that consists of two cascaded sub-networks.
Though both MPU~\cite{mpu} and our methods produce multi-resolution point clouds, our method is not a kind of progressive upsampling method. 
There are two distinctive differences. 
First, MPU~\cite{mpu} consists of a cascade of sub-networks, whereas our method produces multi-scale outputs in a single network. 
Second, the MPU progressively trains $L$ subnets using $2L+1$ stages, whereas we apply multi-scale supervision signals to each output scale simultaneously, so training can be completed in only one go.
PU-GCN~\cite{pu-gcn} and Dis-PU~\cite{dis-pu} extract local and global features to learn fine-grained details using graphical convolutional network and attention mechanism. 
However, their method has a high computation demand due to intensive use of $k$NN and self-attention operation.
Our method uses a hierarchical network architecture that is computationally efficient to extract multi-scale features to grasp the fine-grained patterns.

\noindent\textbf{Multi-scale feature representation and aggregation.}
Multi-scale feature representation and aggregation are one of the main problems in visual perception tasks. 
Lin et al. \cite{fpn} proposes a top-down architecture with lateral connections to fuse multi-scale features instead of directly using the pyramidal feature hierarchy for prediction.
Taking one step further, Liu et al. \cite{panet} adds a bottom-up pathway to enhance the entire feature hierarchy. 
Ghiasi et al. \cite{nas-fpn} adopts a neural architecture search to discover a new feature pyramid architecture.
Tan et al. \cite{efficient-det} revises the architecture design to be more intuitive and principled.
The main difference between our work and the pyramid feature architecture in 2D visual perception tasks is that the latter obtains the multi-scale feature directly from feature extractors. 
But our approach generates a pyramidal point cloud feature from a bi-directional up/downsampling pathway with shortcut connections.

\section{APPROACH}
\subsection{Overview}
\label{sec:network_overview}
Given a sparse point cloud $\mathcal{P}$ of $N$ points, our network outputs L denser point cloud $\{\hat{Q}_{1},\ldots,\hat{Q}_{l},\ldots, \hat{Q}_{L}\}$ where $\hat{Q}_{l} \in \mathbb{R}^{r'N\times 3}$, $r'=2^{l}$ is the intermediate upsampling factor and $r=2^{L}$ is the desired upsampling factor. $Q$ is the ground truth point cloud with $rN$ points.
The upsampled point clouds should lie on the underlying surface of the object and have a uniform distribution. 
Our network consists of three parts: point feature extraction, point feature expansion and point reconstruction.

\noindent\textbf{Point feature extraction.} 
The feature extractor learns point feature $F \in \mathbb{R}^{N\times C_{1}}$ from the input point cloud $\mathcal{P} \in \mathbb{R}^{N\times d}$, $C_1 > d$. 
In this case, the feature of the input is the 3D coordinates of the point cloud, namely $d=3$.
We adopt the feature extractor in MPU~\cite{mpu} which uses dynamic graph convolution~\cite{dynamic-gcn} to extract point features from local neighborhoods via $k$NN search in feature space and exploit a dense connection to facilitate information reuse.

\noindent\textbf{Point feature expansion.} 
In this part, we propose a bi-directional multi-scale upsampling module to expand the point features. 
It takes the point feature $F$ as input and outputs $L$ upsampled point features $\{F^{up}_{1},\ldots,F^{up}_{l},\ldots, F^{up}_{L}\}$, where $F^{up}_{l} \in \mathbb{R}^{r'N\times C_{2}}$, $F^{up}_{L} \in \mathbb{R}^{rN\times C_{2}}$ and $C_2<C_1$.
The bi-directional up and downsampling path preserves the fine-grained geometric details by learning the up/down-sampling operators with global and local context provided by the point features of different resolutions.

\noindent\textbf{Point reconstruction.} 
To reconstruct the $L$ upsampled point features from latent space to coordinate space, we assign each upsampled feature with one 2-layer Multi-Layer Perceptron (MLP) to regress the 3D coordinates.

\subsection{Bi-directional Multi-scale Upsampling Module}
\begin{figure}[t]
\begin{center}
  \includegraphics[width=\linewidth]{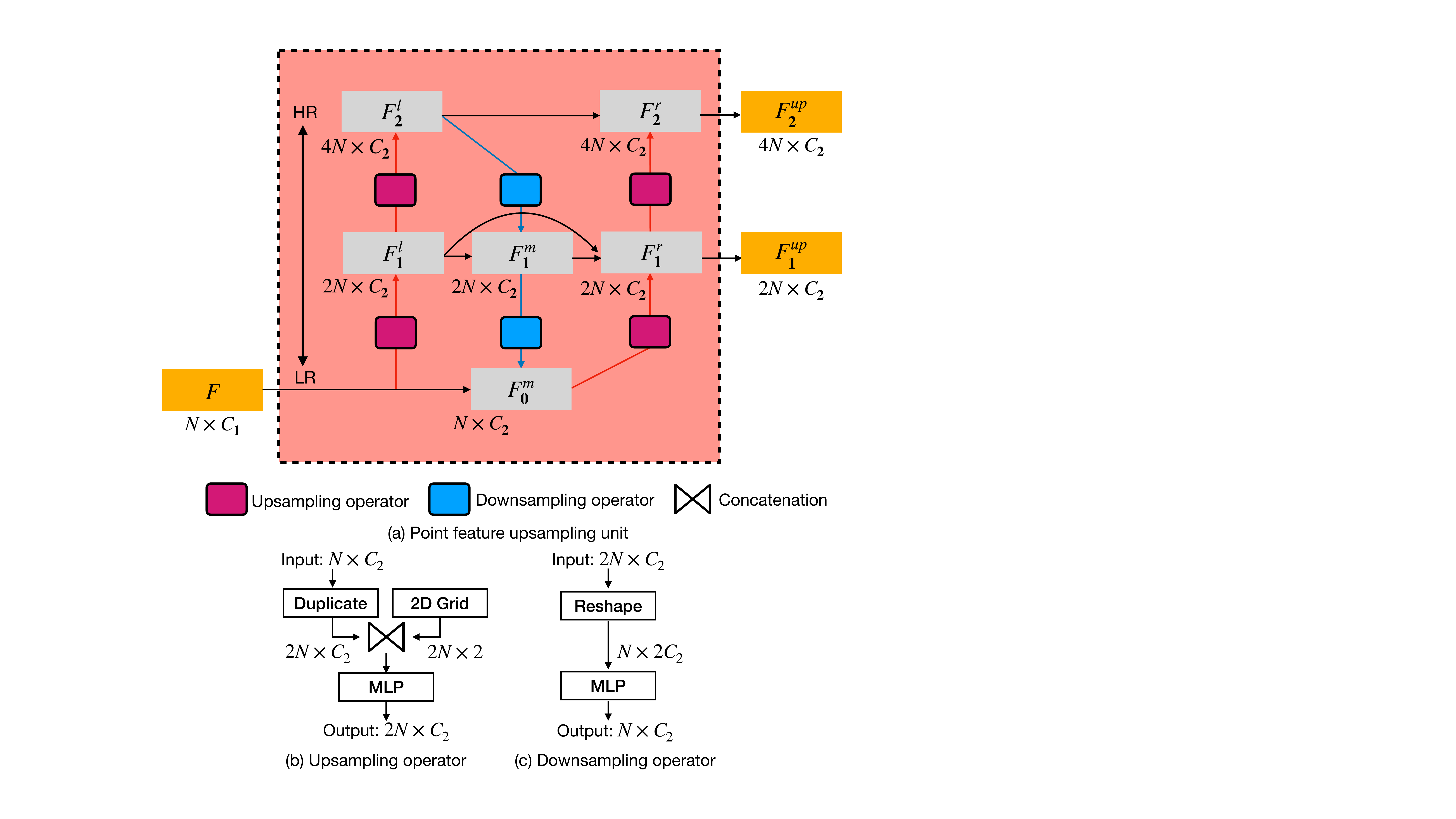}
\end{center}\vspace{-3mm}
  \caption{
  Illustration of the point feature upsampling unit in Fig. \ref{fig:network_overview}. 
  For simplicity, the target upsampling factor is 4.}
  \vspace{-8mm}
\label{fig:upsampling_module}
\end{figure}

The proposed bi-directional multi-scale upsampling module consists of two sets of upsampling operators and one set of downsampling operators as shown in Fig~\ref{fig:upsampling_module}(a). 
The number of operator in each set is $L$, where $2^L=r$. 
The scaling factor of each operator is 2.
We illustrate the up/downsampling operator in Fig~\ref{fig:upsampling_module}(b)/(c).
This upsampling operator combines the duplicated point features and a 2D grid~\cite{foldingnet, mpu} that adds spatial variation to help spread out the points. 
Then it uses shared MLP to produce the upsampled point features.
The downsampling operator reshapes the feature and then uses shared MLP to generate the downsampled point features.
We leverage hierarchical network architecture which is computationally efficient to learn local/global fine grained patterns instead of using self-attention operator or graphical convolutional network.
Given point feature $F$ from feature extraction as input, upsampling module first maps the low-resolution (LR) point feature to a high-resolution (HR) point feature using the first set of upsampling operators on the left side.
Then, it maps the HR point feature back to the LR point feature using the set of downsampling operators in the middle. 
Lastly, the reconstructed LR point feature is mapped to a HR feature point by the second set of upsampling operators on the right side.
Our upsampling module uses the shortcut connections and a weighted feature fusion method to achieve a fast and efficient multi-scale feature fusion.
For simplicity the desired upsampling factor $r$ is set to 4. Concretely, we describe the two upsampled point feature outputs $F_{1}^{\text{up}}, F_{2}^{\text{up}}$, and an intermediate fused feature $F_{1}^{m}$ in Fig.~\ref{fig:upsampling_module}.
% ####################################################################
% ####################################################################
\begin{equation}
\begin{aligned}
\small
F_{2}^{\text{up}} &=\frac{w_{1} \cdot F_{2}^{l}+w_{2} \cdot \operatorname{Up}\left(F_{1}^{r}\right)}{w_{1}+w_{2}+\epsilon} \\
F_{1}^{\text {up}} &=\frac{w'_{1} \cdot F_{1}^{l}+w'_{2} \cdot F_{1}^{m}+w'_{3} \cdot \operatorname{Up}\left(F_{0}^{m}\right)}{w'_{1}+w'_{2}+w'_{3}+\epsilon} \\
F_{1}^{m} &=\frac{w''_{1} \cdot F_{1}^{l}+w''_{2} \cdot \operatorname{Down}\left(F_{2}^{l}\right)}{w''_{1}+w''_{2}+\epsilon} \\
\end{aligned}
\end{equation}

\noindent where $\operatorname{Up} \text{and} \operatorname{Down}$ is the up and downsampling operation; $w_i$ is a learnable weight that represents the importance of each input, which is ensured to be positive by applying $\operatorname{ReLU}$~\cite{relu} after each $w_i$; $\epsilon$ is a small value to avoid numerical instability.

\subsection{Residual Block} % Find a better name !!
\begin{figure}[t]
\begin{center}
% \vspace{-6mm}
  \includegraphics[width=\linewidth]{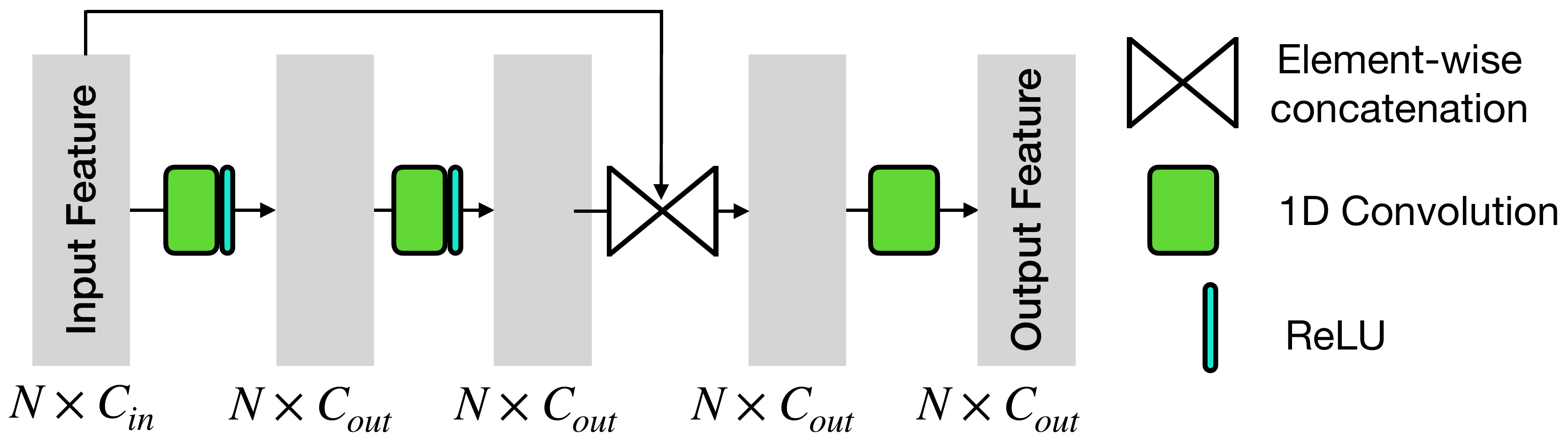}
\end{center}\vspace{-3mm}
  \caption{Illustration of the residual block.}
  \vspace{-6mm}
\label{fig:residual_block}
\end{figure}

In this work, we design a lightweight residual block, as illustrated in Fig.~\ref{fig:residual_block}, to expand/squeeze the channel of point feature during feature expansion and point reconstruction. 
Residual learning have proven to be effective at easing the optimization of a network during training~\cite{resnet}. 
Our intention is to use the residual block to allow the network to learn a complex mapping with increased model complexity namely the scalable bidirectional pathway and cross-scale shortcut links in our model.
% ####################################################################
% ####################################################################
\subsection{Multi-scale Supervision.} 
Our model generates multiple intermediate upsampled point clouds in \emph{one} feed-forward pass. 
Supervision signals are applied \emph{simultaneously} at each output scale. 
Notably, we do not downsample the ground truth to create the multi-scale supervision label to avoid potential artificial artifacts and laborious effort. 
The multi-scale supervision guides the network's training to fuse multi-scale features to be more \emph{discriminative}.
Additionally, decomposing the task into multiple simpler sub-tasks decrease the optimization difficulty. 
It allows our model to have a large representation capacity to learn complicated mappings, thus achieve fewer outliers and render better geometric details.
We train our upsampling network with multi-scale supervision using a robust joint loss:
\begin{equation}
\begin{aligned}
\mathcal{L} &= \sum_{i=1}^{L}\alpha_{i}\cdot\mathcal{L}_{\text{joint}}\left(Q, \hat{Q}_{i}\right) \\
\mathcal{L}_{\text{joint}} &= \mathcal{L}_{\text{CD}} + \lambda \cdot \mathcal{L}_{\text{rep}}
\end{aligned}
\label{eqn:joint_loss}
\end{equation}

\noindent where $Q$ and $\hat{Q}_i$ are the ground truth and multi-scale output point cloud respectively; $\alpha_{i}, \lambda \in [0,1]$ are weighting factors; $\mathcal{L}_{\text{CD}} \left(\cdot \right)$ means Chamfer Distance \cite{point_generation} which measures the average closest point distance between two point sets; $\mathcal{L}_{\text{rep}} \left(\cdot \right)$ means Repulsion Loss \cite{pu-net} which encourage the generated points to distribute more uniformly.

\section{EXPERIMENTS}
\subsection{Implementation Details}
\label{sec:implementation_details}
\noindent\textbf{Datasets}
For quantitative/qualitative comparisons between models, we employ two synthetic datasets and one real-scanned dataset. 
(1) PU-GAN's dataset provides 120 training and 27 testing objects.
(2) PU1K dataset proposed in PU-GCN~\cite{pu-gcn} consists of 1,020 training and 127 testing objects.
% ###############################################################################
\begin{figure*}[t]
\begin{center}
  \includegraphics[width=\linewidth]{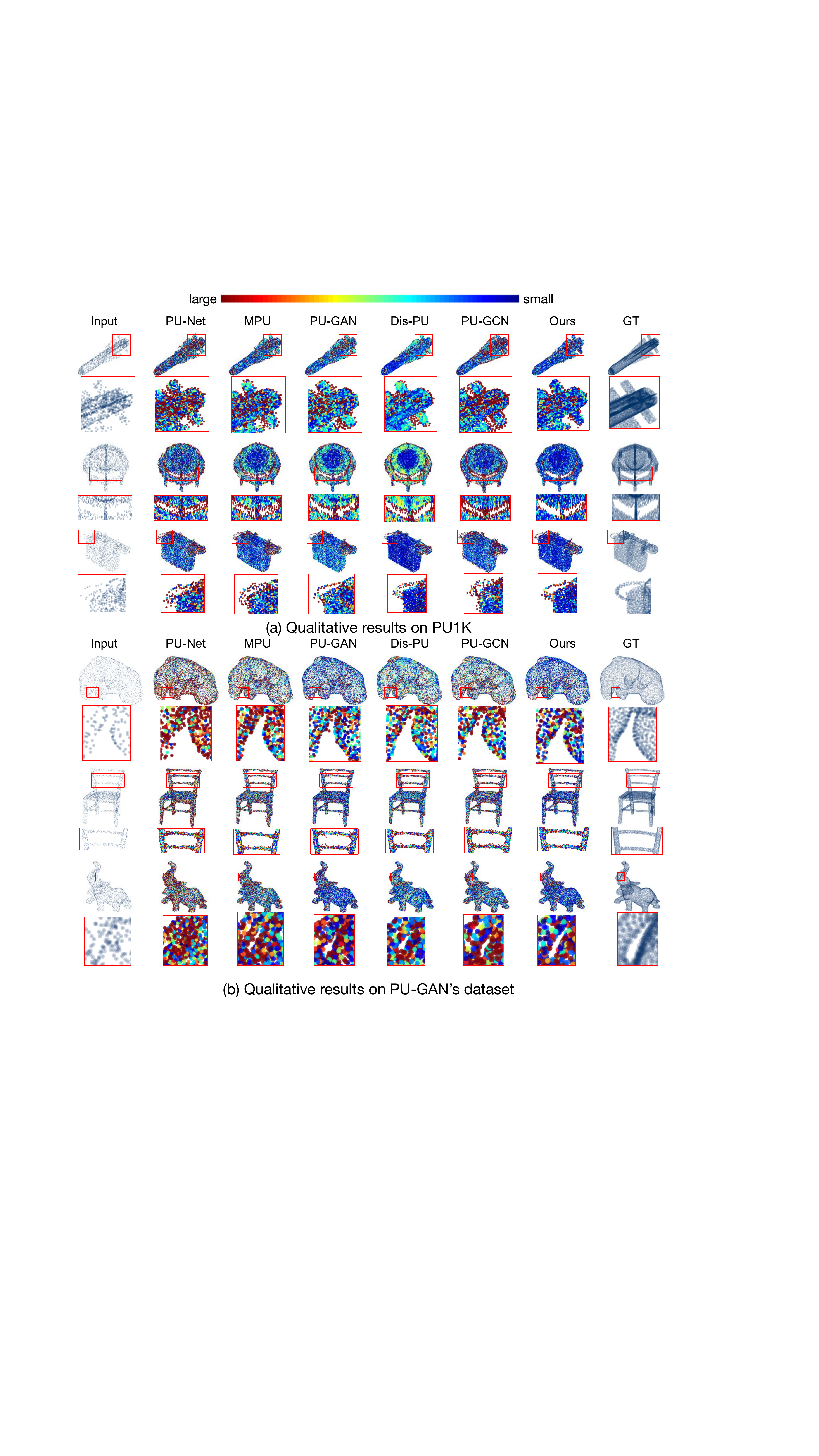} 
\end{center} \vspace{-3mm}
  \caption{Qualitative comparisons of point cloud upsampling on synthetic dataset.
%   The objects in the first two rows are from PU1K~\cite{pu-gcn}, while objects in the last two rows are from PU-GAN~\cite{pu-gan}'s dataset.
  The color indicates the nearest distance of each output point to the ground truth surface. 
  We can see that our model produces a more accurate reconstruction with fewer red points and preserves better geometric details at challenging areas.}
  \vspace{-6mm}
  \label{fig:qualitative_synthetic}
% \vspace{-4mm}
\end{figure*}
% ###############################################################################
(3) The real-scanned dataset ScanObjectNN~\cite{ScanObjectNN} contains 2,902 point cloud objects in 15 categories manually filtered and selected from SceneNN~\cite{scenenn} and ScanNet~\cite{scan-net}. Each object has 2,048 points.
Since the ground truth is not available, we only conduct qualitative experiments on the real-world dataset with pre-trained models.
To demonstrate point cloud upsampling is beneficial to 3D object classification, we employ both synthetic and real-scanned datasets in our experiment in Section~\ref{sec:benefit_to_classification}.
The synthetic dataset is the ModelNet40~\cite{shapenet} dataset, which contains point clouds of 40 common object categories sampled from 100 unique CAD models per category.
The real-scanned dataset is ScanObjectNN~\cite{ScanObjectNN}.

\noindent\textbf{Training details}
For training, we use the training data provided by PU-GAN's dataset~\cite{pu-gan} and PU1K~\cite{pu-gcn} and follows the settings in PU-GAN~\cite{pu-gan} and PU-GCN~\cite{pu-gcn} for model comparison. 
Concretely, the ground truth point clouds $Q$ has 1,024 points; 
the input point clouds $P$ of 256 points are randomly downsampled from the ground truth point cloud $Q$ on the fly during training; the upsampling ratio $r$ is 4.
We train our model for 400 epochs using the Adam optimizer with an initial learning rate of 0.001.
The batch size is 64 on PU1K and 28 on PU-GAN's dataset.
We decrease the learning rate by a factor of 0.7 for every 40 epochs. 
Data augmentation techniques applied includes random rotation/scaling/shifting.
In the point feature extraction unit the $k$ for k-nearest neighbors search is 16.
$C_1$ and $C_2$ in the feature expansion unit are 648 and 128.
We use $\alpha_1=0.6, \alpha_2=1.0$ and $\alpha_1=0.6, \alpha_2=0.8, \alpha_3=1.0$ in Eq.~\ref{eqn:joint_loss} for $r=4$ and $r=16$ respectively. 
We implemented our network using PyTorch and all experiments are conducted on an Nvidia RTX 2080 GPU.

\noindent\textbf{Testing details}
For testing, we adopt the commonly used patch-based strategy~\cite{pu-net, mpu, pu-gan, pu-gcn, dis-pu} as follows.
First, use Poisson disk sampling to generate the ground-truth object point clouds $Q$ of $rN$ points from object mesh and then downsample it to get sparse input point clouds $P$ of $N$ points.
Second, apply the farthest point sampling~\cite{pointnet++} to the input point clouds to get query points and extract overlapping input patches of 256 points around each query point using $k$NN.
Next, feed all input patches to an upsampling model and combine the output overlapping point clouds of 1,024 points to get the dense object point cloud.
Lastly, apply farthest point sampling to produce a uniform and dense object point cloud that contains $rN$ points.
% Highlight the test setting differences.
Both PU-GCN and Dis-PU reported model comparisons results using PU-GAN's dataset. 
We notice two differences in test settings between them and PU-GAN's in their experiment.
(1) PU-GAN~\cite{pu-gan} and Dis-PU~\cite{dis-pu} use Monte-Carlo downsampling while PU-GCN uses Poisson downsampling\footnote{https://github.com/guochengqian/PU-GCN/issues/3\#issuecomment-888289259} to generate the sparse input point cloud $P$.
(2) PU-GAN~\cite{pu-gan} and PU-GCN~\cite{pu-gcn} use input point cloud of 2,048 points but Dis-PU~\cite{dis-pu} uses input point cloud of 1,024 points.
Because the Monte-Carlo downsampling generates a realistic and non-uniform point cloud distribution, while the Poisson downsampling produces a uniform point cloud distribution,
and the former is also used during training for input point cloud generation.
Hence, we follow PU-GAN~\cite{pu-gan} to use the widely used Monte-Carlo sampling.
Regarding the number of test input points, we also follow PU-GAN~\cite{pu-gan}'s setting to use 2,048 points for consistency between model comparisons conducted on PU-GAN's dataset and PU1K dataset.

\noindent\textbf{Evaluation metrics}
The evaluation metrics are (i) Chamfer distance (CD); (ii) Hausdorff distance (HD)~\cite{hd}; (iii) point-to-surface distance (P2F). 
A lower evaluation metric indicates a better performance.

\subsection{Quantitative Comparisons}
\label{sec:quantitative_comparisons}
We conduct comparisons on two datasets PU-GAN's dataset~\cite{pu-gan} and PU1K~\cite{pu-gcn}. 
The results are shown in Tables~\ref{tab:benchmark_pugan} and~\ref{tab:benchmark_pu1k}, respectively. 
The recently proposed PUGeo-Net~\cite{pugeo} is not included in the comparison as it requires the accurate normal of point for training, which is not directly available in point clouds.
% ###############################################################################
\begin{figure*}[t]
\begin{center}
  \includegraphics[width=\linewidth]{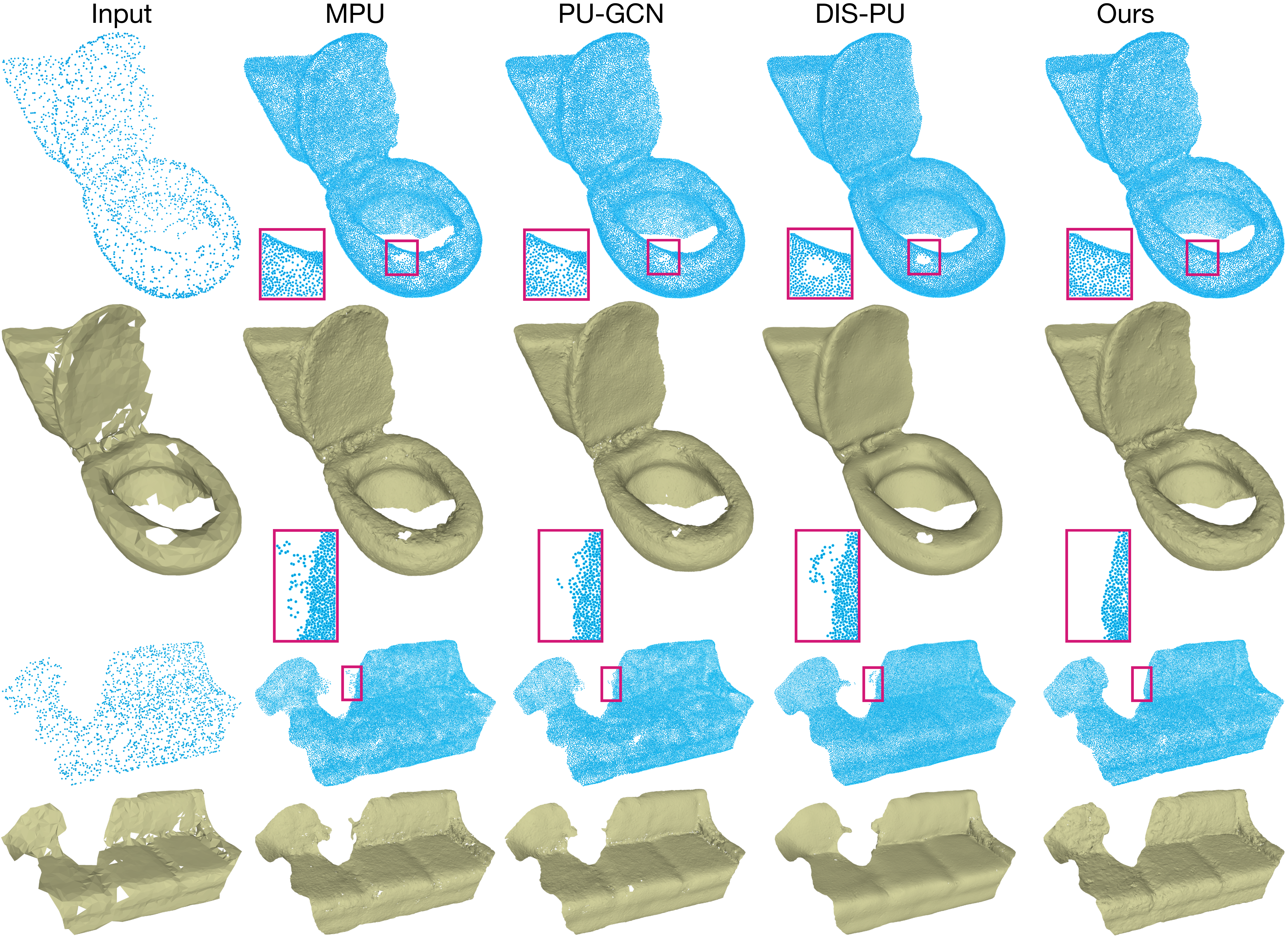} 
\end{center}
\vspace{-3mm}
  \caption{
  Qualitative comparison of upsampled (16$\times$) point clouds generated using different methods and its 3D mesh reconstruction. 
  The sparse inputs of 2,048 points are from real-scanned dataset ScanObjectNN~\cite{ScanObjectNN}.
  While Dis-PU~\cite{dis-pu} has an advantage in surface smoothness, our method outperforms others in terms of producing a more accurate reconstruction that has fewer surface defects and outliers.}
  \label{fig:qualitative_real}
\vspace{-6mm}
\end{figure*}
% ###############################################################################

\begin{table}[t]
	\centering
	\caption{
    \bd{Quantitative comparisons with the state-of-the-art on PU-GAN's dataset.}
    The units of CD, HD, and P2F are $10^{-3}$. 
    The best result is highlighted in bold letters and the runner-up is highlighted with an underline.
	}
	\label{tab:benchmark_pugan}\vspace{-1mm}
	%	\begin{center}
	\resizebox{\linewidth}{!}{
		\begin{tabular}{@{\hspace{1mm}}c@{\hspace{1mm}}||
				@{\hspace{1mm}}c@{\hspace{1mm}}||c@{\hspace{3mm}}c@{\hspace{3mm}}c@{\hspace{1mm}} ||
				@{\hspace{1mm}}c@{\hspace{1mm}}||c@{\hspace{3mm}}c@{\hspace{3mm}}c@{\hspace{1mm}}
			} \shline
			\multirow{2}*{Methods}
			& \multicolumn{4}{@{\hspace{1mm}}c@{\hspace{1mm}}||@{\hspace{1mm}}}{4$\times$}
			& \multicolumn{4}{@{\hspace{1mm}}c@{\hspace{1mm}}@{\hspace{1mm}}}{16$\times$} \\
			
			\cline{2-5} \cline{6-9}
			& Size & CD$\downarrow$ & HD$\downarrow$ & P2F$\downarrow$ & Size & CD$\downarrow$ & HD$\downarrow$ & P2F$\downarrow$\\ \hline \hline
			PU-Net~\cite{pu-net}
			&10.1M & 0.72  &  8.94  &  6.84
			&24.5M & 0.38  & 6.36  & 8.44      \\
			MPU~\cite{mpu}
			&23.1M & 0.49  & 6.11  & 3.96
			&92.5M & 0.19  & 5.58  & 3.52      \\
			PU-GAN~\cite{pu-gan}
			& 9.6M & 0.28  & 4.64  & 2.33
			& 9.6M & 0.23  & 6.09  & 3.31      \\
			PU-GCN~\cite{pu-gcn}
			& 9.7M & \underline{0.27}  &  \underline{4.38}  & 2.80
			& 9.7M & \underline{0.18}  &  \underline{4.72}  & 3.15      \\
			Dis-PU~\cite{dis-pu}
			&13.2M & \textbf{0.24} & 4.63  & \underline{2.23}
			&13.2M & \textbf{0.16} & 8.14  & \textbf{2.43}    \\ 
			\hline
			Our
			& 8.3M & 0.28 & \textbf{4.28} & \textbf{2.05}
			&15.6M & \textbf{0.16} & \textbf{4.70} & \underline{2.59}
			\\ \shline
	\end{tabular}}
	%\end{center}
	\vspace*{-7mm}
\end{table}
\noindent\textbf{Comparisons on PU-GAN's dataset.} 
Quantitative comparisons between models on PU-GAN's dataset under different upsampling ratios are presented in Table~\ref{tab:benchmark_pugan}.
The results show that our method performs competitively to state-of-the-art approaches under both small and large upsampling ratios.
Notably, our model has the smallest model size when $r=4$ and is 37\% lighter than the runner-up model Dis-PU.
Though our model size grows as the upsampling ratio increases, when $r=16$, it is still comparable to the size of Dis-PU and much smaller than PU-Net and MPU.
The input and ground truth test data are generated following PU-GAN's setting when testing $r=4$ and $r=16$.
We train the models on PU-GAN~\cite{pu-gan}'s dataset using their released source code and report the test performance using the best model obtained from training.
For $r=4$, we directly use the result of PU-Net, MPU, and PU-GAN from PU-GAN~\cite{pu-gan}'s paper.
Though Dis-PU~\cite{dis-pu} and PU-GCN~\cite{pu-gcn} have conducted model comparisons using PU-GAN's dataset under $r=4$ and $r=16$, we don't use the results in their paper because they used different test settings from PU-GAN~\cite{pu-gan}, which is discussed in Section~\ref{sec:implementation_details}-Testing in detail.

\noindent\textbf{Comparisons on PU1K.} 
Quantitative comparisons on PU1K dataset are presented in Table \ref{tab:benchmark_pu1k}.
We conduct two sets of experiments. 
One uses test input point cloud generated using Poisson downsampling provided by PU-GCN~\cite{pu-gcn}.
The other one uses input point cloud generated using Monte-Carlo downsampling to produce a more realistic and non-uniform distribution.
For PU-Net~\cite{pu-net}, MPU~\cite{mpu}, and PU-GCN~\cite{pu-gcn}, we use the pre-trained model provided by PU-GCN~\cite{pu-gcn} to get their results.
\begin{table}[t] \centering
  \tablestyle{2.5pt}{1.0} 
  \caption{\textbf{Quantitative comparisons with the state-of-the-art on PU1K.}
  We conduct two sets of experiments. 
  One uses input point cloud generated using Poisson downsampling as in the paper~\cite{pu-gcn}.
  Another one uses input point cloud generated using Monte-Carlo downsampling which produces more realistic non-uniform distribution distribution.
  The units of CD, HD, and P2F are $10^{-3}$. 
  The best result is highlighted in bold letters and the runner-up is highlighted with an underline.}
  \vspace{-2mm}
  \label{tab:benchmark_pu1k}
\begin{tabular}{c|c|c|c|c|c|c|c}
\shline
\multirow{2}{*}{Model} & \multirow{2}{*}{Size} & \multicolumn{3}{c|}{Monte-Carlo} & \multicolumn{3}{c}{Poisson}\\
\cline{3-8}

                        &               &CD$\downarrow$     &HD$\downarrow$   &P2F$\downarrow$    &CD$\downarrow$    &HD$\downarrow$    &P2F$\downarrow$\\
\hline
 PU-Net \cite{pu-net}  &10.1M            &0.623             &10.907           &2.877              &1.155             &15.170            &4.834 \\
 MPU \cite{mpu}        &23.1M            &0.534             &9.725            &2.286              &0.935             &13.327            &3.551 \\
 PU-GAN \cite{pu-gan}  &\underline{9.6M} &0.439             &\underline{7.697}&2.117              &0.727             &9.622             &2.936 \\
 PU-GCN \cite{pu-gcn}  & 9.7M            &0.462             &7.671            &2.125              &0.585             &\underline{7.577} &2.499 \\
 Dis-PU \cite{dis-pu}  &13.2M            &\underline{0.423} &\textbf{7.095}   &\underline{1.645}  &\underline{0.550} &\textbf{6.997}    &\textbf{2.240} \\
\hline
 Our                   &\textbf{8.3M}    &\textbf{0.420}    &7.990            &\textbf{1.547}     &\textbf{0.541}    &8.360             &\underline{2.280} \\
 \shline
\end{tabular}
\vspace{-8mm}
\end{table}
As the pre-trained model of PU-GAN~\cite{pu-gan} and Dis-PU~\cite{dis-pu} is not provided, we train the models on PU1K using their released codes and get the result using the best model.
In both sets of experiments, our model presents competitive performance.
We can see that all models generally performs better when the input point cloud is generated using Monte-Carlo downsampling as it is the downsampling method used during training (see Section~\ref{sec:implementation_details}-Training details).
Our superior performances on the Monte-Carlo setting verified that our model is more robust to non-uniform points.
We also get lower errors on CD and P2F, which shows that our model is able to reconstruct more accurate object shapes.
\subsection{Ablation Study}
\begin{table}[t] \centering
\tablestyle{3.5pt}{1.0} 
% \vspace{-5mm}
\caption{
\bd{Ablation study.} 
Experiments are conducted on PU-GAN's dataset. 
The units of CD, HD and P2F are $10^{-3}$. 
MS is the abbreviation of multi-scale.
The effectiveness of our proposed components (residual block, multi-scale supervision, and multi-scale feature fusion) for point cloud upsampling is validated.
}
\label{tab:ablation}\vspace{-1mm}
\begin{tabular}{c|c|c|c|l|l|l}
\shline
Model       & \footnotesize{Residual}    & \footnotesize{MS Supervision} & \footnotesize{MS Fusion}  & CD$\downarrow$  & HD$\downarrow$ & P2F$\downarrow$\\ 
\shline
A           &                                  &                               &              & 0.30 & 6.05 & 2.77      \\
B           & \checkmark                       &                               &              & 0.28 & 5.24 & 2.17      \\
C           & \checkmark                       & \checkmark                    &              & 0.28 & 4.87 & 2.04      \\
\hline
Full        & \checkmark                       & \checkmark                    & \checkmark   & 0.28 & 4.28 & 2.05      \\
\shline
\end{tabular}
\vspace{-6mm}
\end{table}

We analyze the contribution of each component of our network on the PU-GAN's dataset in Table \ref{tab:ablation}, which includes multi-scale fusion, multi-scale supervision, and residual block. 
We remove each component from the full model one by one (from bottom to top) and measure the performances in terms of CD, HD and P2F. As shown in Table \ref{tab:ablation}, all components contribute to the full model since removing any component hampers the performance.

\subsection{Qualitative Comparisons}
We compare our model qualitatively with other methods on two synthetic datasets and one real-word dataset.
The results are shown in Fig.~\ref{fig:qualitative_synthetic} and~\ref{fig:qualitative_real}.
In Fig.~\ref{fig:qualitative_synthetic}, the color indicates the nearest distance of each output point to the ground truth surface.
We observe three distinct advantages of our approach:
1) Generate points with lower error in the area near the sharp edge, and the edges are cleaner and sharper.
2) Produces fewer outliers in challenging areas like the joint, intersection, and the narrow gap between two surfaces.
3) Preserve better geometric details of slender objects.
Specifically, the number of points generally increases along the longitudinal direction of the objects.
In Fig.~\ref{fig:qualitative_real}, we compare the upsampled point cloud generated using different methods and its 3D mesh reconstruction, the noisy and sparse inputs are from the real-scanned dataset ScanObjectNN~\cite{ScanObjectNN}, where we set $r=16$.
While Dis-PU~\cite{dis-pu} has an advantage in surface smoothness, our method outperforms others in terms of producing a more accurate reconstruction that has fewer surface defects and outliers.
The qualitative results suggest that our model possesses a better understanding of the global and local context relationship and is capable of generating high-fidelity object details.

\subsection{Benefit of upsampling to point cloud classification}
\label{sec:benefit_to_classification}
\begin{table}
\begin{center}
\tablestyle{7pt}{1.0}
\caption{
\bd{Classification accuracy comparison.} 
We use random sampling and farthest point sampling on the testing point cloud in synthetic dataset ModelNet40~\cite{shapenet} and real-scanned dataset ScanObjectNN~\cite{ScanObjectNN} to generate point clouds of 128 points as input and point clouds of 512 points as ground truth.
Then we upsample the input point cloud by 4 times using a set of upsampling models.
The classification accuracy comparison is conducted between input/upsampled/ground-truth point cloud.
The results indicate that point cloud upsampling is beneficial to point cloud classification, and our model is superior in generating new points that reflect the underlying surface of point clouds.
Overall and average class accuracy are shown in \%.
}
\label{tab:classification_comparison}\vspace{-1mm}
\begin{tabular}{c|c|c|c|c}
\shline
\multirow{2}{*}{Model} & \multicolumn{2}{c}{ModelNet40~\cite{shapenet}} & \multicolumn{2}{c}{ScanObjectNN~\cite{ScanObjectNN}} \\
                        & OA. & Cls Acc.  & OA & Cls Acc. \\
\hline
PU-GAN \cite{pu-gan}    & 84.2      & 80.0      & 71.5      & 67.4\\
PU-GCN \cite{pu-gcn}    & 84.6      & 80.6      & 70.9      & 67.0\\
Dis-PU \cite{dis-pu}    & 85.0      & 80.4      & 71.1      & 66.8\\	
Ours                    & \textbf{87.1}      & \textbf{82.5}      & \textbf{72.0}      & \textbf{68.2}\\
\hline
Input                   & 79.4      & 74.1      & 61.8      & 56.3\\
Ground truth            & 92.5      & 88.7      & 74.9      & 70.6\\
\shline
\end{tabular}
\end{center}
\vspace{-9mm}
\end{table}
Point cloud object classification is a fundamental task to robot perception which is crucial to downstream tasks like object detection and semantic segmentation.
To demonstrate the value of point cloud upsampling to the robotic community, we design an experiment to show that upsampling is beneficial to point cloud object classification.
In this experiment, 
We employ a synthetic dataset ModelNet40~\cite{shapenet} and a real-scanned dataset ScanObejctNN~\cite{ScanObjectNN}
and choose two widely used models PointNet++\footnote{https://github.com/yanx27/Pointnet\_Pointnet2\_pytorch}~\cite{pointnet++} and PointNet\footnote{https://github.com/hkust-vgd/scanobjectnn}~\cite{pointnet} to perform point cloud shape classification on ModelNet40 and ScanObjectNN respectively.
The PointNet++ is pre-trained on ModelNet40, whereas PointNet is pre-trained on ScanObjectNN's hardest variant PB\_T50\_RS.
First, we use farthest point sampling and random sampling to sample the testing point clouds to point clouds of 512 points as ground truth data and point clouds of 128 points as input data.
Next, we upsample the input point clouds by four times using a set of point cloud upsampling models.
The upsampling models are pre-trained on PU-GAN's~\cite{pu-gan} dataset used in Table~\ref{tab:benchmark_pugan}.
We compare the classification accuracy of the ground-truth, randomly-downsampled, and upsampled point clouds in Table~\ref{tab:classification_comparison}.
The results show that applying point cloud upsampling to sparse and nonuniform point clouds to generate denser point clouds effectively improves the classification result in both synthetic and real-scanned data.
Interestingly the classification performance improvement is more significant on real-scanned data.
This demonstrates that the upsampling models can generate new points according to the distribution pattern of the underlying surface.
Further, the quantitative classification accuracy comparison indicates the superiority of our model in reconstruction accuracy.
Experiments could be designed following a similar principle to show the effect of point cloud upsampling on semantic segmentation and part segmentation.

\section{CONCLUSION}
In this work, we propose a bi-directional multi-scale upsampling approach for 3D point cloud upsampling. 
We decompose a bi-directional up/downsampling pathway into sub-up/downsampling steps of smaller scaling factors to produce a pyramidal multi-scale point feature hierarchy.
The point features in the hierarchy are fused and reconstructed to point clouds of different resolutions. 
Supervision signals are applied to each output point cloud to ensure that the feature fusion produces discriminative features.
A simple yet effective residual block is proposed to reduce the optimization difficulty.
Extensive quantitative and qualitative results on synthetic and real-world datasets demonstrate that our method achieves superior results compared to state-of-the-art approaches.
We demonstrate that point cloud upsampling can improve robot perception by ameliorating the 3D data quality using a simple experiment.

%%%%%%%%%%%%%%%%%%%%%%%%%%%%%%%%%%%%%%%%%%%%%%%%%%%%%%%%%%%%%%%%%%%%%%%%%%%%%%%%
\bibliographystyle{./IEEEtran} % use IEEEtran.bst style
\bibliography{./IEEEabrv,./ral_icra_2022}

\end{document}